\documentclass[sigconf]{acmart}
\usepackage{balance}
\usepackage{xspace}
\usepackage[utf8]{inputenc}
\usepackage{multirow}

\newcommand*{\etal}{\emph{et al}.\@\xspace}

\begin{document}


\title{An Occam's Razor View on Learning Audiovisual Emotion Recognition with Small Training Sets}

\author{Valentin Vielzeuf}
\affiliation{%
    \institution{Orange Labs}
  \city{Cesson-S\'evign\'e} 
  \country{France} 
  \postcode{35512}
}
\additionalaffiliation{%
    \institution{Normandie Univ., UNICAEN, ENSICAEN, CNRS}
  \city{Caen} 
  \state{France} 
}
\email{valentin.vielzeuf@orange.com}

\author{Corentin Kervadec}
\affiliation{%
    \institution{Orange Labs}
  \city{Cesson-S\'evign\'e} 
  \country{France} 
  \postcode{35512}
}
\email{corentin.kervadec@orange.com}

\author{Stéphane Pateux}
\affiliation{%
    \institution{Orange Labs}
  \city{Cesson-S\'evign\'e} 
  \country{France} 
  \postcode{35512}
}
\email{stephane.pateux@orange.com}

\author{Alexis Lechervy}
\affiliation{%
    \institution{Normandie Univ., UNICAEN, ENSICAEN, CNRS}
  \city{Caen} 
  \country{France} 
}
\email{alexis.lechervy@unicaen.fr}

\author{Frédéric Jurie}
\affiliation{%
    \institution{Normandie Univ., UNICAEN, ENSICAEN, CNRS}
  \city{Caen} 
  \country{France}
}
\email{frederic.jurie@unicaen.fr}

\begin{abstract}
This paper presents a light-weight and accurate deep neural model for audiovisual emotion recognition.
To design this model, the authors followed a philosophy of simplicity, drastically limiting the number of  parameters to learn from the target datasets, always choosing the simplest learning methods: i) transfer learning and  low-dimensional space embedding allows to  reduce the dimensionality of the representations. ii) The visual temporal information is handled by a simple score-per-frame selection process, averaged across time. iii) A simple frame selection mechanism is also proposed to weight the images of a sequence. iv) The  fusion of the different modalities is performed at prediction level (late fusion).
We also highlight the inherent challenges of the AFEW dataset and the difficulty of model selection with as few as 383 validation sequences.
The proposed real-time emotion classifier achieved a state-of-the-art accuracy of 60.64~\% on the test set of AFEW, and ranked 4th at the Emotion in the Wild 2018 challenge.
\end{abstract}

\ccsdesc[500]{Computing methodologies~Computer vision tasks}
\ccsdesc[500]{Computing methodologies~Neural networks}
\ccsdesc[300]{Computing methodologies~Computer vision representations}

\keywords{Emotion Recognition; Deep Learning;}

\maketitle
\section{Introduction}
Emotion recognition is a current topic of interest, finding many applications such as health care, customer analysis or even face animation. With the advance of deep learning for face analysis, automatic emotion recognition might appear as an already solved problem. 
Indeed, large image datasets for facial expression recognition in uncontrolled conditions are emerging, allowing to learn accurate deep models on this kind of task. For instance, EmotioNet~\cite{benitez2017emotionet} gathers one million faces annotated in Action Units~\cite{ekman1977facial},  AffectNet~\cite{mollahosseini2017affectnet} proposes half a million usable faces annotated in both discrete emotion~\cite{plutchik2013theories} and arousal valence~\cite{barrett1999structure}, and Real-World Affective Faces~\cite{li2017reliable} is a dataset of around  30,000 faces with very reliable and accurate annotations of the discrete and compound emotions.

However, the algorithms proposed in the literature do not allow yet to reach a human-level understanding of the emotions. It is the reason why several multimodal and temporal datasets have been proposed recently. The AVEC challenge~\cite{ringeval2017avec} contains a dataset of videos annotated with per-frame arousal valence and several features. The FERA challenge~\cite{valstar2017fera} presents a corpus of different head poses to allows action units detection in uncontrolled temporal conditions. Finally AFEW~\cite{dhall2012collecting,dhall2015video} is annotations in discrete emotions of 773 training audiovisual clips extracted from movies. These clips are therefore very noisy, due to the uncontrolled conditions. 

Nevertheless these datasets contain a small amount of samples and therefore raise three issues for machine learning approaches: i) how to cope with the temporal aspect of emotions? ii) how to combine the modalities? iii) how to learn a meaningful representation from so few samples?
We investigate these issues by focusing on the AFEW dataset annotated with discrete emotion.
Several methods have been exploring these questions during the past years. 
The literature associated with the first challenge editions focuses on hand-crafted features~\cite{yao2015capturing,kachele2016revisiting} (e.g.  Local Binary Patterns, Gabor filters, Modulation Spectrum, Enhanced AutoCorrelation, Action Units), which are then fed to classifiers such as SVM or Random Forest. After 2015, visual learned features are becoming the dominating approach, with the use of large deep convolutional neural networks~\cite{fan2016video}. To handle the problem of the small size of these datasets, recent approaches~\cite{knyazev2018leveraging,hu2017learning,vielzeuf2017temporal,fan2016video} use transfer learning from models learned on larger image datasets. To handle the temporal nature of the signal, several authors use LSTM recurrent neural networks~\cite{gers1999learning,hu2017learning,vielzeuf2017temporal,fan2016video} even if no strong improvements have been obtained, compared to other simpler methods, as observed by Knyazev~\etal~\cite{knyazev2018leveraging}. Other authors propose to use 3d convolutions \cite{fan2016video} and, possibly combined with LSTM~\cite{vielzeuf2017temporal}.
The audio modality is often described by hand-crafted features on top of which a classifier is trained, even in recent approaches\cite{hu2017learning,knyazev2018leveraging,vielzeuf2017temporal}. Pini~\etal\cite{pini2017modeling} also propose to use a Soundnet~\cite{aytar2016soundnet} as a features extractor. 
Finally, to combine the modalities, late fusion approaches are preferred and performs better on this dataset~\cite{vielzeuf2017temporal}.

We present in this paper a light-weight real-time neural network model based on an "Occam's razor" philosophy, consisting in always choosing the simplest method at equal performance. The methods section details the characteristics of this model, while results section reports its performance on several visual benchmarks and on the EmotiW audiovisual challenge~\cite{AFEW18}.
\section{Methods}
\label{methods}
This section presents the proposed framework for audiovisual emotion recognition, by discussing first the visual and audio modalities, and then by proposing a method for fusing them. We finally address the issues raised by the (small) size of the dataset.
\begin{figure}[ht]
\centering
\includegraphics[width=0.9\linewidth]{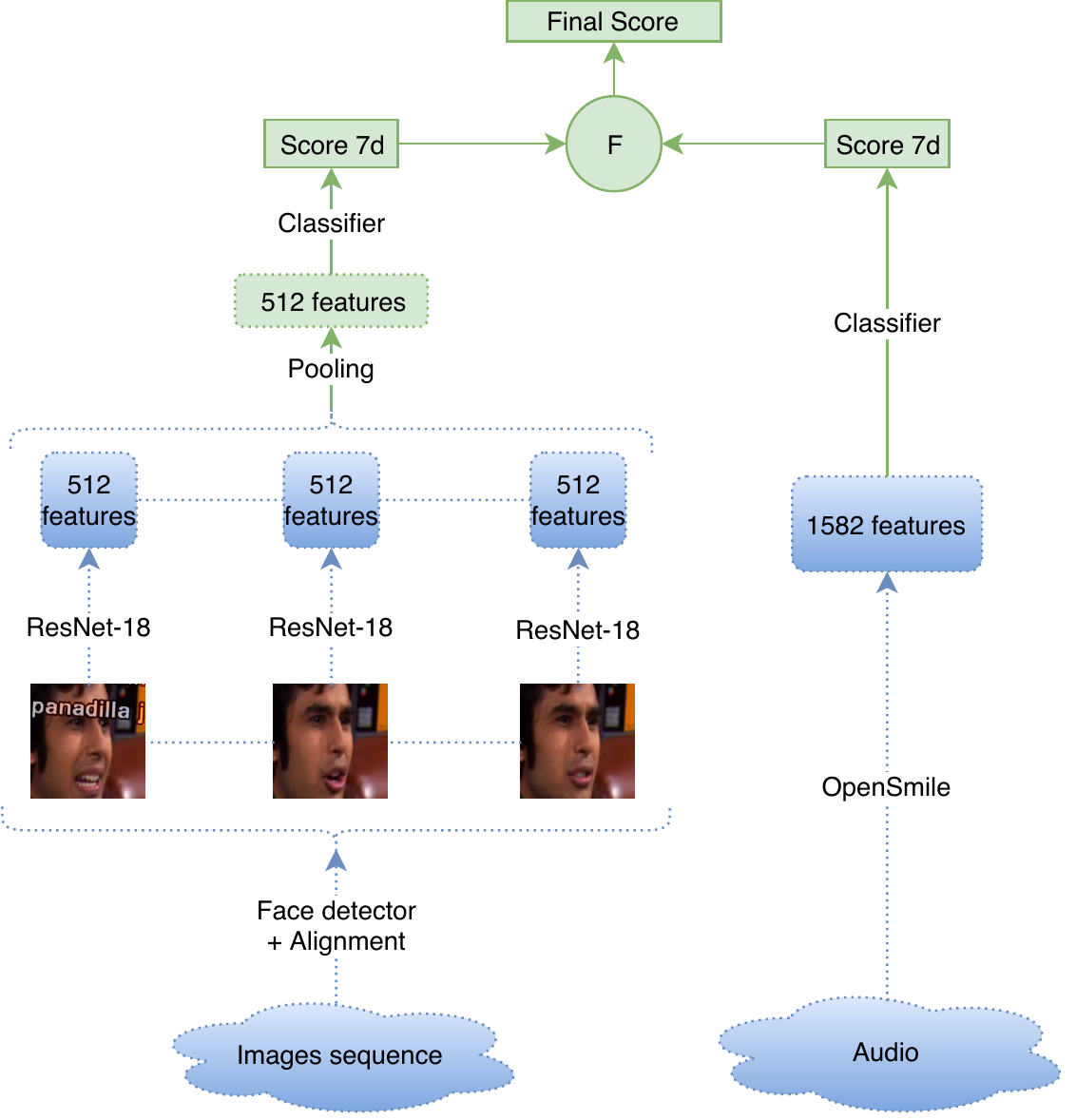}
\caption{Overview of our framework. Only the green part (on the right side) is trained on AFEW.}
\label{overview}
\end{figure}

\subsection{Visual modality}
The visual modality consists in sequences of images extracted from the AFEW videos. They are processed by: i) applying a face detector ii) aligning the landmarks with a landmark detector and an affine transform,  resizing aligned faces to $224\times 224$. The so-obtained faces are referred as the {\em visual} input in the rest of the paper.

\paragraph{Emotion classification with a ResNet-18 network} 

Our first objective is to learn a still image  emotion classifier, based on the ResNet-18~\cite{he2016deep}, which we considered to be well adapted to this task.  However, the small amount of training clips (773) makes it hard to learn from scratch. Consequently, we first trained it on the larger AffectNet dataset~\cite{mollahosseini2017affectnet}, containing in the order of 300,000 usable faces, annotated with both emotion labels (8 labels, not the same as AFEW) and arousal valence values.  Multi-task learning allows us to use these two types of annotations, by  replacing the last dense layer of the ResNet-18 network by two dense layers: one arousal-valence linear regressor and one emotion classifier (softmax layer). Their two losses are optimized during training, leading to a more general 512-sized hidden representation. 
We also use standard regularization methods such as data augmentation (e.g. jittering, rotation) to be robust to small alignment errors, cutout~\cite{devries2017improved} and dropout\cite{srivastava2014dropout}.

Applying directly this model to AFEW face images gives not better than chance results, due to the big differences in the annotations of the 2 datasets. To deal with this issue, we use two other emotion datasets: i) Real-World Affective Faces (RAF)~\cite{li2017reliable}, containing 12,271 training images annotated with emotion labels, with a large number of annotators for each sample and hence a high confidence in the annotations. ii) SFEW~\cite{dhall2011static}, containing fewer than 1,000 training images, but extracted from AFEW's frames and annotated with the same labels. The fine-tuning is done on these two datasets, in two times. In a first time, only the parameters of the last layer (regressor/classfier) are optimized. On a second time, all the parameters of the network are fine-tuned, but with a lower learning rate.
The so-obtained classifier can be applied to the AFEW faces, giving, for each frame, the arousal-valence prediction and class label scores. We also output, for further use, the weights of the hidden layer (512 weights), used as face features.

\paragraph{From still image to video classification}
Once processed by the above explained model, and assuming $L$ denotes the number of frames of a video, we obtain a set of $L$ 512-dimensional face descriptors with their associated scores (one score for each one of the seven categories) as well as  arousal/valence predictions. 

One very simple way to aggregate the temporal information, and produce classification scores at the level of the video, is to average the per-frame scores. As shown in the experiment sections, it already gives very good results (see Section~\ref{sec-without-aggreg}). However, we propose to explore better ways in the following.

\begin{figure}[tb]
\centering
\includegraphics[width=0.7\linewidth]{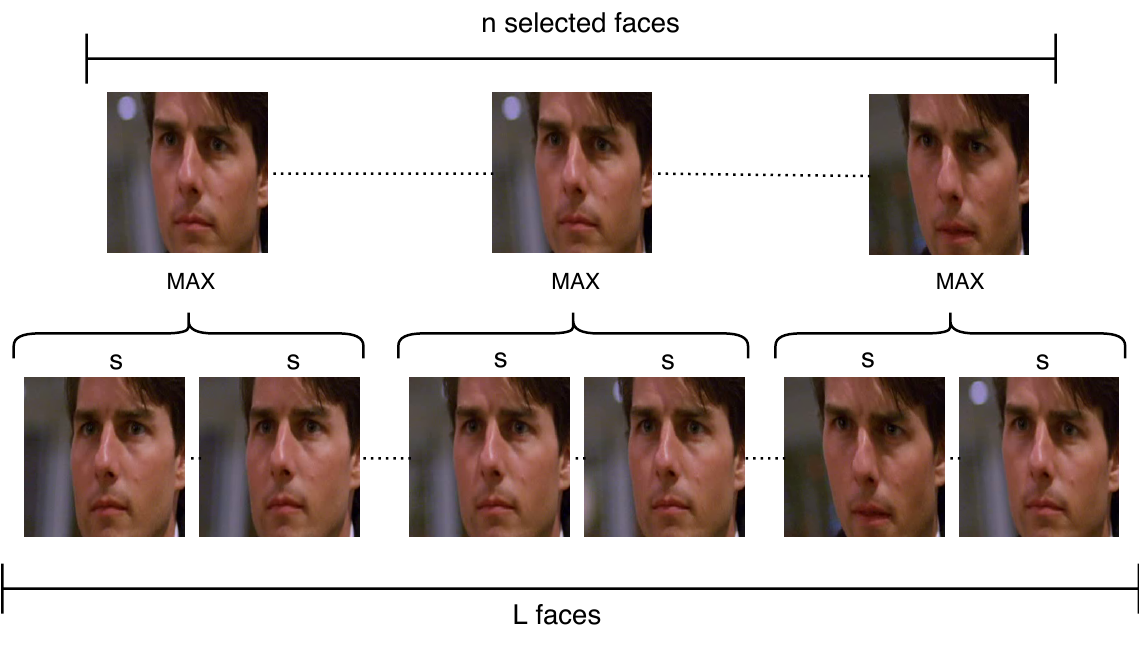}
\caption{$n$ faces are selected from a sequence of $L$ frames. $s$ is the score of a given frame.}
\label{selection}
\end{figure}

We select $n$ faces ($n=16$ in the rest of the paper) from the $L$ original ones, combining down-sampling and max-pooling, as shown in Figure~\ref{selection}. We first divide the sequence into $n$ chunks of equal length, and choose in each chunk the face with the highest score (across frames and categories), represented by $512-d$ feature vectors. For the rest, the per-frame scores are not used anymore.

At the end of this stage, we have a $nx512$ tensor representing the faces of the video sequence.
The final video clip classification is obtained by temporal pooling of the $n$ face features. We consider in the experiment two alternatives: the first one consists in a simple average of the face features, the second one consists in a weighted average of these features. In both cases, once the features are averaged, it gives a 512-d vector on which a linear classifier (softmax with cross entropy) is applied. 
The $n$ weights of the weighted average, one per selected frame, are regressed with a linear regressor (with sigmoid activation) applied to the arousal-valence representation of the frames. The regressor is trained jointly with the whole network. 
The third fusion method consists in training a LSTM network with 128 hidden units on the 512-d features of the $n$ selected frames. The output of the LSTM is the scores of the 7 classes.

Please note that for the sake of simplicity and because we observed no improvement, we do not compute temporal features (such as C3D features), contrarily to several recent approaches~\cite{vielzeuf2017temporal,fan2016video}.

\subsection{Audio modality}
Regarding the audio modality, we experimented two alternatives.

The first one consists in extracting 1582-d features designed for emotion recognition, using the OpenSmile toolbox~\cite{eyben2010opensmile} trained on the IEMOCAP dataset~\cite{busso2008iemocap}.
On top of these features, we  apply a random forest classifier~\cite{breiman2001random}.

The second one consists in extracting the same OpenSmile features and train a fully connected classifier with 64 hidden units, batch-normalization, dropout, and reLu activation. The so-obtained model is then fine-tuned on the AFEW train set.

\subsection{Fusion of the modalities}
Image and audio give each one a set of scores (with one score per class).
As shown by previous works\cite{vielzeuf2017temporal}, sophisticated fusion methods often performs worse than simpler ones on AFEW, probably because of their large number of parameters and the small size of the dataset. We therefore experimented two simple approaches for computing the final scores:
(a) the mean of the modality scores;
(b) the weighted mean of the score, learned on the validation set, similar to the last year's winning approaches~\cite{fan2016video,hu2017learning}.

\subsection{Ensemble learning}
Ensemble learning is often used to boost the results, as observed in last competition works~\cite{zeng2017crafting,miech2017learnable,kim2016hierarchical,antipov2016apparent,yu2015image}. We implement an ensemble of our temporal model, by learning several times the same model with different initializations, and averaging their  predictions.

\subsection{Dealing with the small size of AFEW}

The proposed method has several hyper parameters or architectures details (size of layers, etc.) that have to be set. The most common way to set these parameters is to choose those giving the best performance on the validation set. However, due to the small size of the training and validation sets of AFEW, this is rather unreliable. We experimented with 3 alternatives :  (i) training several times the model with different initialization and computing the mean performance and the standard deviation (std); (ii) merging the training and validation sets and applying cross-validation; (iii) using estimated per-class accuracy and weight classes according to the test set distribution.
The third alternative, as previously done by \cite{knyazev2018leveraging,vielzeuf2017temporal}, uses the distributions given in Table~\ref{distrib}. The accuracy  on the weighted validation set is then computed as follows: $a_{pond} = \sum_{i=1}^{7} a_{i} \frac{n_{i}}{653}$, where $a_{i}$ the estimated accuracy of the $i^{th}$ class on the validation set, $n_{i}$ the number of elements of this class in the test set and $653$ the number of samples of the test set.

\begin{table}[tb]
\centering
\begin{tabular}{lllllllll}
\hline
      & \multicolumn{1}{c}{An.} & \multicolumn{1}{c}{Di.} & \multicolumn{1}{c}{Fe.} & \multicolumn{1}{c}{Ha.} & \multicolumn{1}{c}{Sa.} & \multicolumn{1}{c}{Ne.} & \multicolumn{1}{c}{Su.} & \multicolumn{1}{c}{All} \\ \hline
Train & 133                       & 74                          & 81                       & 150                       & 117                     & 144                         & 74                           & 773                       \\
Val   & 64                        & 40                          & 46                       & 63                        & 61                      & 63                          & 46                           & 383                       \\
Test  & 99                        & 40                          & 70                       & 144                       & 80                      & 191                         & 29                           & 653                       \\ \hline
\end{tabular}
\caption{AFEW dataset: number of video sequences per class.}
\label{distrib}
\end{table}

\section{Results}
This section experimentally validates the proposed model, by first presenting the results obtained by our audio and visual models taken individually, and, combined in a second time for the audiovisual challenge. It is worth noting that our model uses a relatively small number of parameters and can work in a real-time setting (180M FLOPs). Our performance is measured by training several models (in the order of 50 models) with different initialization and measuring the mean accuracy and standard deviation (std).

\subsection{Emotion Recognition in Images\label{sec-without-aggreg}}
We first compare our ResNet-18 model pre-trained on AffectNet, without temporal aggregation of the features, to several state-of-the-art methods, on emotion  classification in images. As shown in Table~\ref{visual}, despite its small number of parameters, our model gives very good results, outperforming its competitors on the AFEW validation set. In this case, temporal fusion is done by averaging per frame predictions. Note that we measure a std around 0.5\%.

\begin{table}[tb]
\centering
\begin{tabular}{llllll}
\hline
Model                & RAF  & SFEW  & AFEW  & FLOP    & Param         \\ \hline
CNN Ensemble~\cite{yu2015image}      & \_   & 55.96 & \_    & \textgreater 2000 & \textgreater 500 \\
HoloNet~\cite{hu2017learning}              & \_   & \_    & 46.5  & 75                &  \_                   \\
Cov. Pooling~\cite{acharya2018covariance}   & \textbf{85.4} & \textbf{58.14} & 46.71 & 1600              & 7.5              \\
Transfer VGG~\cite{vielzeuf2017temporal}         & \_   & 45.2  & 41.4  & 1550              & 138              \\
Our (image) & 80   & 55.8  & \textbf{49.4}  & 180      & \textbf{2}                \\ \hline
\end{tabular}
\caption{Accuracy of different models for facial expression classification. Weights and FLOPs are in millions. Transfer VGG is a VGG-face model fine-tuned on FER-2013~\cite{goodfellow2013challenges}.}
\label{visual}
\end{table}

\subsection{Evaluation of the Temporal Pooling} 
We provide here the evaluation of the 3 different temporal pooling methods we proposed in Section~\ref{methods}. Performance is measured as the accuracy on the validation set of AFEW (mean and std). We also indicate the weighted accuracy (see details in Section~\ref{methods}), which can be seen as a more accurate estimation of the actual performance on the test set. The standard deviation is, on average, around 0.6\%. 

Our main observations are: i) temporal features aggregation is useful,  ii) average and weighted average have approximately the same performance and are better than our LSTM model. iii) combining weighted and non-weighted average pooling seems to help a little on the weighted validation set. iv) our best performance outperforms any results published yet on this dataset. However, with a std of 0.6\%, we must be cautious about the conclusions. 

 We also note that the std on the weighted validation set is a bit lower (resp. 0.5\% and 0.4\% for av. pool. and weighted av. pooling). It can be explained by higher variations of performance inside the "difficult" classes (disgust, surprise), which are rare in the test set. 

\begin{table}[tb]
\centering
\begin{tabular}{clcc}
\hline
\multicolumn{1}{l}{}     & Model                      & Accuracy             & Weig. Acc.    \\ \hline
\multirow{10}{*}{\rotatebox[origin=c]{90}{Visual}} & no feat. aggregation            & 49.4                   & 55.6          \\
                         & av. pool. (1)                  & 49.7                   & 60.5          \\
                         & av. pool. (4)  & 50.4                   & 61.2          \\
                         & av. pool. (50) & {52.2}          & 61.7          \\ \cline{2-4} 
                         & weig. av. pool   (1)               & 50.2                   & 61.1          \\
                         & weig. av. pool (4)  & 50.3                   & 61.5          \\
                         & av. pool (2) + weig. av. pool (2)  & 50.1                   & {62.0} \\ \cline{2-4} 
                         & LSTM 128 hidden units      & 49.5                   & 58.2          \\
                         & \textit{VGG-LSTM}~\cite{vielzeuf2017temporal}          & {48.6}          & \_            \\
                         & \textit{FR-Net-B}~\cite{knyazev2018leveraging}          & {{53.5}} & \_            \\ \hline\hline
\multirow{3}{*}{\rotatebox[origin=c]{90}{Audio}}   & MLP                        & 33.5                   & 42.1          \\
                         & MLP  (pre-trained)          & 35.0                     & 45.2          \\
                         & Random Forest              & 38.8                   & 44.3          \\ \hline
\end{tabular}
\caption{Accuracy for visual and audio modalities and for state-of-the-art visual models (italic). Indication '($x$)' means: ensemble of $x$ models with different initialization}.
\label{temporalaudio}
\end{table}

\subsection{Evaluation of the Audio Modality}
As for the visual modality, we evaluate the 3 proposed methods. The std is of 1.5\% for the Multi Layer Perceptron trained from scratch and of 0.8\% for the Multi Layer Perceptron  pre-trained on IEMOCAP. 
Even if the Random Forest yields good results, it over-fits the training set with the accuracy of almost 100\% with hence a high risk of bad generalizations. The pre-trained MLP achieves the best and most stable results. 
The audio modality is weaker than the visual one and the AFEW annotation of a video seems to rely more on the visual modality and on the context. Nevertheless, audio can bring a $+3\%$ accuracy gain when combined with visual modality, as reported by Fan~\etal\cite{fan2016video}, and is not an option for this challenge.

\subsection{Audiovisual Challenge}
This section describes our 7 submissions to the 2018 edition of the EmotioW challenge. For each submission, we explain the audio and the visual modalities in Table~\ref{challenge}, as well as the performance.

First submission: most simple model with average pooling of visual features and MLP for audio. Video and audio scores are weighted resp. with 0.65 and 0.35 (weights learned on the validation set).
Second submission: audio/video weights set to 0.5/0.5 and combination of RF and MLP.
Third submission (our best one): ensemble of 6 models (see Table~\ref{challenge} for details).
Fourth submission:  same as the third submission, but replacing the Random Forest by a second MLP.
Fifth and sixth submissions: larger visual ensembles.
Seventh submission: same as third one using Train+Val for training (surprisingly 0.1\% lower).

\begin{table}[tb]
\begin{tabular}{lllll}
\hline
\#& Visual                                                           & Audio                                                & Weigh. Val & Test \\ \hline
1          & av. pool.  (1)                                                       & MLP (1)                                                 & 62.1      & 57.2 \\ \hline
2          & av. pool. (1)                                                         & \begin{tabular}[c]{@{}l@{}}RF (1)\\  +MLP (1)\end{tabular} & 62.4      & 58.6 \\ \hline
3          & \begin{tabular}[c]{@{}l@{}}av. pool. (2)\\ + weig. av. pool.(2)\end{tabular} & \begin{tabular}[c]{@{}l@{}}RF (1)\\ +MLP (1)\end{tabular} & 62.7      & 60.6 \\ \hline
4          & weig. av. pool. (2)                                                      & MLP (2)                                                & 63.5      & 59.4 \\ \hline
5          & \begin{tabular}[c]{@{}l@{}}av. pool. (2)\\ + weig. av. pool.(2)\end{tabular} & MLP (2)                                                & 63.0      & 60.4 \\ \hline
6          & av. pool. (50)                                                      & MLP (2)                                                & 63.6      & 59.4 \\ \hline
7          & av. pool. (4)                                                       & MLP (1)                                                & 72.4      & 60.5 \\ \hline
\end{tabular}
\caption{Our 7 seven submissions to the 2018 Emotion in the Wild challenge. Indication '($x$)' means: ensemble of $x$ models trained with different initialization. See text for details. }
\label{challenge}
\end{table}
These results confirmed our three intuitions.
First, performance on validation and test sets are very different, making it difficult to choose a model from the validation set accuracy. Second, adding the validation set to the train set for final training makes the performance worse, which is counter-intuitive. Last but not least, drastically limiting the number of trainable parameters on such a small dataset is one of the key ingredients to better generalizations. 

These experiments also highlights the importance of having a weighting scheme to improve the selection of the model on the validation set, knowing that  validation and test sets have different distributions. More generally, we also observed large standard deviations in our cross-validation experiments, explaining why it's difficult to compare different methods on the validation set. 

On overall, the proposed light model is real time  and achieved the accuracy of 60.64\%, allowing it to ranked 4th at the 2018 edition of the EmotiW Challenge. 

We do believe that the performance on this challenge starts saturating, which can be explained by the small size of the dataset and the subjective (and therefore noisy) nature of the annotations. We indeed noted that human performance reported on the validation set of AFEW by ~\cite{vielzeuf2017temporal} is comparable to the performance reached by our model. There is consequently a risk that improving the performance on this dataset will  consist in exploiting its biases rather than actually learning a better representation of emotions.
\section{Conclusions}
This paper proposes a new audiovisual model for emotion classification in videos. This model is carefully designed following the "Occam's razor" principle, which can be summed up by "always choose the simplest approach". For both modalities we limited the number of trainable parameters to their minimum. Transfer learning is also used to include reliable a priori knowledge and solve the high-dimensional versus lack of data paradigm, especially for the visual modality. A basic but well-performing temporal pooling is also proposed, including a frame selection mechanism. Finally, a simple fusion method average of the score limits again the number of  parameters of the model.

\balance
\bibliographystyle{ACM-Reference-Format}
\bibliography{bibliography} 

\end{document}